\documentclass{article}

 \usepackage[preprint]{neurips_2025}

\usepackage[utf8]{inputenc} % allow utf-8 input
\usepackage[T1]{fontenc}    % use 8-bit T1 fonts
\usepackage{hyperref}       % hyperlinks
\usepackage{url}            % simple URL typesetting
\usepackage{booktabs}       % professional-quality tables
\usepackage{amsfonts}       % blackboard math symbols
\usepackage{nicefrac}       % compact symbols for 1/2, etc.
\usepackage{microtype}      % microtypography
\usepackage{xcolor}         % colors
\usepackage{amsmath}
\usepackage{graphicx}

\title{MMRPT: MultiModal Reinforcement Pre-Training via Masked Vision-Dependent Reasoning}

\author{
  Xuhui Zheng\thanks{Equal contribution.}~~\thanks{Word done during internship at SenseTime.} \\
  SenseTime, Nanjing University\\
  \texttt{zhengxuhui@smail.nju.edu.cn} \\
  \And
  Kang An$^{*\dagger}$ \\
  SenseTime, Shenzhen University \\
  \texttt{ankang@gml.ac.cn} \\
  \And
  Ziliang Wang$^{*}$ \\
  SenseTime \\
  \texttt{wangziliang1@sensetime.com} \\
  \And
  Yuhang Wang \\
  SenseTime \\
  \texttt{wangyuhang@sensetime.com} \\
  \And
  Faqiang Qian \\
  SenseTime \\
  \texttt{qianfaqiang@sensetime.com} \\
  \And
  Yichao Wu\thanks{Corresponding author} \\
  SenseTime \\
  \texttt{wuyichao@sensetime.com}
}

\begin{document}

\maketitle

\begin{abstract}
% Multimodal pre-training remains heavily limited by the descriptive bias and coverage constraints of image–caption pairs, causing models to rely on surface linguistic patterns rather than grounded visual understanding. We address this limitation with MMRPT, a masked multimodal reinforcement pre-training framework that enhances deep visual reasoning in MLLMs. We automatically construct masked multimodal data by estimating each sentence’s visual dependency-computed via attention over visual tokens-and masking highly vision-dependent segments. The model must reconstruct these segments through vision-grounded reasoning, guided by a reward model that evaluates semantic correctness and visual faithfulness. Empirically, MMRPT provides uniform zero-shot gains and markedly improves fine-tuning stability, while theoretically demonstrating that reinforcing vision-grounded masked reasoning during pre-training can fundamentally strengthen the model’s cross-modal abstraction and generalization mechanisms.

Multimodal pre-training remains constrained by the descriptive bias of image–caption pairs, leading models to favor surface linguistic cues over grounded visual understanding. We introduce \textbf{MMRPT}, a masked multimodal reinforcement pre-training framework that strengthens visual reasoning in MLLMs. \textbf{We are the first to incorporate reinforcement learning directly into the pre-training of large vision–language models}, enabling learning signals that reward visual grounding rather than caption imitation. MMRPT constructs masked multimodal data by estimating sentence-level visual dependency via attention over visual tokens and masking highly vision-dependent segments; the model reconstructs these spans through vision-grounded reasoning guided by a semantic–visual reward. Experiments show consistent zero-shot gains across diverse benchmarks and substantially improved robustness under supervised fine-tuning, demonstrating that reinforcement-driven masked reasoning provides a more reliable and generalizable pre-training objective for multimodal models.

\end{abstract}

\section{Introduction}
Multimodal large language models (MLLMs) have rapidly advanced the integration of vision and language\cite{yin2024survey, ghosh2024exploring,jin2024efficient, caffagni2024revolution,chen2024lion}, achieving strong performance on tasks such as image captioning, visual question answering, and instruction following. These models typically rely on large-scale image–text corpora\cite{schuhmann2022laion}, where the dominant training paradigm is to align visual features with descriptive captions and optimize autoregressive objectives\cite{liu2023visualinstructiontuning, chen2024internvl, bai2025qwen2,vteam2025glm45vglm41vthinkingversatilemultimodal,yu2025minicpmv45cookingefficient,zhang2025pelicanvl10foundationbrain} by next token predictions or contrastive learning\cite{radford2021learning,yu2022coca,coreteam2025mimovltechnicalreport}. While effective for learning general vision–language correspondences, this paradigm exhibits a fundamental limitation: the model’s \textbf{visual understanding capacity is bounded by the descriptive quality, granularity, and bias of captions}\cite{sim2025can,zhang2024visually,vo2025vision,zhao2025looking,li2021align, lewis2024does}. Captions often provide partial, object-centric, or stylistically constrained descriptions, leaving substantial portions of the visual scene underexplored.

Consequently, MLLMs trained in this manner tend to adopt \textit{text-first strategies}\cite{zhao2025looking,park2025second}, relying on learned linguistic priors or dataset biases rather than deeply grounding their predictions in visual evidence\cite{lin2023revisiting,zhao2025mitigate,zheng2025modality}. This leads to well-documented issues such as visual hallucination, shallow cross-modal alignment, and limited abstraction abilities when reasoning about fine-grained or implicit visual content\cite{bai2024hallucination,chen2024multi,dai2025see}. As the field pushes toward human-level visual reasoning\cite{zhou2025perception,islam2024large}, these limitations reveal a growing mismatch between current pre-training pipelines and the cognitive processes required for genuine visual understanding\cite{park2025second,zheng2024thinking}.

A key question therefore arises: \textbf{How can we move beyond imitation-driven, caption-limited learning and enable MLLMs to acquire deeper, inference-driven visual reasoning capabilities during pre-training?} A solution must not only reduce the over-reliance on captions but also provide a scalable mechanism for encouraging models to actively infer visual information, rather than passively mimic textual patterns.

To address these challenges, we propose \textbf{MMRPT}, a \textit{Masked Multimodal Reinforcement Pre-Training framework} that rethinks how visual understanding should be acquired during multimodal pre-training. Instead of learning passively from captions, MMRPT encourages models to \textit{actively infer} missing visual information by embedding reinforcement learning directly into the pre-training pipeline. This shift transforms multimodal pre-training from a descriptive, imitation-driven process into a reasoning-centered paradigm.

The key idea behind MMRPT is to leverage the model’s own decoding behavior to identify which parts of the text genuinely depend on visual evidence. We estimate a \textbf{visual dependency score} for each sentence by analyzing attention distributions over visual tokens in selected Transformer layers. Sentences identified as highly vision-dependent are then masked, forming a large-scale corpus of \textit{vision-sensitive masked reasoning tasks} constructed automatically from ordinary image–text data. Unlike traditional masked modeling, where missing tokens can often be reconstructed through language priors, MMRPT ensures that the masked segments require \textbf{vision-grounded reasoning}, pushing the model to engage with the image rather than rely on textual co-occurrence.

A reinforcement learning objective further strengthens this process. A reward model evaluates both the semantic correctness and visual faithfulness of reconstructed segments, guiding the policy toward deeper perceptual reasoning during the later stages of pre-training. This mechanism allows MMRPT to scale naturally with existing multimodal corpora, without requiring curated annotations or synthetic reasoning traces.

Through this design, MMRPT introduces three conceptual advances to multimodal pre-training:
\begin{itemize}
    \item (1) \textbf{Active visual inference:} enabling models to reason about masked content that truly depends on visual evidence;
    \item (2) \textbf{Attention-driven data construction:} leveraging the model’s own attention patterns to automatically create vision-sensitive training signals;
    \item (3) \textbf{Reinforcement-based semantic grounding:} using reward-driven optimization to cultivate deeper, meta-level visual reasoning abilities beyond imitation learning.
\end{itemize}
Together, these improvements move MLLMs closer to genuine visual understanding, reducing reliance on caption priors and enhancing their ability to interpret implicit, fine-grained, or abstract visual concepts that captions typically omit.
\section{Method}
\subsection{Identifying Vision-Dependent Language Units}
\label{dentifying Vision-Dependent Language Units}
A core challenge in multimodal pre-training is that many caption tokens can be predicted without relying on the image. To ensure our masked RL tasks genuinely require visual reasoning, we identify visually dependent language at two granularities: sentence-level (coarse) and token-level (refined).

\textbf{(1) Sentence-Level Visual Dependency Estimation}: Given an image–text pair $((I, T={s_1,\dots,s_n}))$, where each ($s_i$) is a sentence, we feed the model in autoregressive mode and extract cross-attention matrices from selected Transformer layers. Let:

$(V)$ denote visual tokens after encoding the image
$(T)$ denote text tokens
$(\text{Attn}_\ell(t, x))$ denote the attention weight from text token $(t)$ to token $(x)$ at layer $(\ell)$

For a sentence $(s_i)$, we compute its \textbf{visual dependency score}:

\begin{equation}
    D_{\text{vis}}(s_i)=
    \frac{1}{|s_i||\mathcal{L}|}
    \sum_{\ell\in\mathcal{L}}
    \sum_{t\in s_i}
    \frac{\sum_{v\in V} \text{Attn}_\ell(t,v)}
    {\sum_{x\in V\cup T}\text{Attn}_\ell(t,x)},
\end{equation}

where $(\mathcal{L})$ is a set of attention layers chosen for visual grounding.

A sentence is classified as \textbf{vision-dependent} if:

\begin{equation}
    D_{\text{vis}}(s_i) \ge \tau,
\end{equation}

where $(\tau)$ is an adaptive threshold (e.g., mean + $\gamma$·std over the entire caption).
This formulation avoids token-level noise and captures semantic groups that genuinely require visual grounding.

\textbf{(2) Token-Level Visual-Need Refinement via a Text-Only LLM}:
Even in a highly visual sentence, not all tokens require image evidence.
We therefore refine candidate masks using a \textbf{text-only LLM} $(f_{\text{text}})$.

Given a sentence $(s_i)$, we prompt the LLM to produce a binary label for each token $(w)$:

\begin{equation}
    y_w = f_{\text{text}}(w \mid s_i, \text{instruction}),
\end{equation}

where $(y_w = 1)$ indicates that humans would need the image to determine the meaning of $(w)$.
We keep only tokens satisfying:

\begin{equation}
    y_w = 1 \quad \text{and} \quad D_{\text{vis}}(s_i) \ge \tau.
\end{equation}

This two-step pipeline ensures that mask units are both \textbf{cognitively aligned} (from textual reasoning) and \textbf{model-grounded} (from visual attention).

\subsection{Constructing Vision-Sensitive Masked Data for RL}
Simply masking tokens in captions is insufficient because models may reconstruct them using textual shortcuts or accidental exposure during training. We design a principled construction mechanism to guarantee that each masked sample requires \textbf{vision-dependent inference}, forming high-quality input for RL.

\textbf{(1) Masking Constraints to Prevent Shortcut Learning.}

\hspace{1.0em} a) \textit{Only one masked token per generated RL sample.}
    
For each caption $(T)$, assume we identify maskable positions (${m_1, m_2,\dots, m_k}$).
Instead of masking all at once, we produce \textbf{$k$ independent RL samples}, each masking a \textbf{single} unit:
\begin{equation}
    T^{(j)} = T \setminus {m_j}, \qquad j=1,\dots,k.
\end{equation}
This prevents mutual inference between masked locations.

\hspace{1.0em} \textit{b) Only mask the \textbf{first} occurrence of repeated tokens.}

While sentence-level visual dependency identifies which sentences likely require visual grounding, not every token within such sentences is genuinely vision-dependent. To further refine the masking targets, we employ an external \textbf{text-only LLM} that operates purely on the original caption text without access to the image.

The LLM is instructed to identifies token spans within image captions that exhibit generative visual necessity-i.e., segments that cannot be contextually inferred from linguistic priors alone and require direct visual input for coherent generation.

\hspace{1.0em} \textit{c) Truncating samples to avoid exposure of answers.}

Let ($s_i$) be the sentence containing the masked token.
We define the final RL input as:

\begin{equation}
    T^{(j)}_{\text{RL}} = {s_1,\dots,s_i^{\text{masked}}},
\end{equation}

discarding all subsequent sentences (${s_{i+1},\dots,s_n}$).
This prevents leakage of the unmasked token later in the caption.

\hspace{1.0em} \textit{d) Maintain sequential order for samples derived from the same caption.}

If a caption yields multiple masked samples (${T_{\text{RL}}^{(1)},\dots,T_{\text{RL}}^{(k)}}$),
their internal training order satisfies:
\begin{equation}
    T_{\text{RL}}^{(1)} \prec T_{\text{RL}}^{(2)} \prec \dots \prec T_{\text{RL}}^{(k)}.
\end{equation}

Globally, the dataset is shuffled, but within each caption group, ordering prevents premature exposure to future information.

\textbf{(2) Resulting Dataset Properties}

The final dataset ($\mathcal{D}_{\text{mask}}$) satisfies:
\begin{equation}
    \mathcal{D}_{\text{mask}} =
\bigcup_{i,j} (I_i, T_{i,\text{RL}}^{(j)}, m_{i,j}),
\end{equation}

where each entry forces the model to reconstruct ($m_{i,j}$) \textbf{only from image evidence}, enabling the subsequent reinforcement learning stage to optimize visual reasoning behavior.

\subsection{Masked Multimodal Reinforcement Pre-Training}
After constructing the vision-sensitive masked dataset ($\mathcal{D}_{\text{mask}}$), we train the model to actively infer the masked visual content using \textbf{masked span reasoning}. Unlike standard supervised masked-prediction objectives, we formulate the task as a reinforcement learning (RL) problem \cite{zhang2025pelicanvl10foundationbrain,wang2025stepsearchignitingllmssearch,wang2025eraseimproveerasablereinforcement}, enabling the model to learn deeper visual reasoning behaviors rather than merely fitting textual co-occurrence patterns.

\textbf{(1) Masked Span Reasoning with Structured Reason→Answer Outputs.}

For each masked sample ($(I, T_{\text{mask}}, m)$), where ($m$) is the masked span,
the model generates two structured segments:
\begin{itemize}
    \item \textbf{a reasoning trace}, enclosed in a special tag (e.g., `<think> ... </think>`), and
    \item \textbf{a final answer}, placed inside `<answer> ... </answer>`.
\end{itemize}

This format encourages the model to explicitly articulate its visual reasoning process before committing to an answer. Let the generative trajectory be ($o$), consisting of ($o_{\text{think}}$) and ($o_{\text{answer}}$).

\textbf{(2) RL Objective design.}

The RL objective is:
\begin{equation*}
    J(\theta) = \mathbb{E}_{(I,T,m)\sim\mathcal{D}_{\text{mask}}}
\mathbb{E}_{o\sim\pi_\theta(\cdot|I,T)}
\left[ r(o, m) \right].
\end{equation*}

We design reward (r(o,m)) using \textbf{structure}, \textbf{semantic}, and \textbf{token-level correctness}.

To ensure the model truly infers the masked span using visual evidence, we revise the answer reward to incorporate \textbf{strict prefix matching} between the predicted answer and the ground-truth span, in addition to string-level \textbf{Exact Match(EM)}. The  reward now becomes:

\begin{equation}
r(o,m)=r_{\text{format}}(o)+r_{\text{ans}}(o_{\text{answer}},m).
\end{equation}

\hspace{1.0em} \textit{a) String-level Exact Match (EM).}

As before, EM checks whether the normalized decoded answer exactly matches the ground-truth masked content:
\begin{equation}
    \text{EM}(o_{\text{answer}}, m)=
\begin{cases}
1 & \text{if normalized strings are identical}, \\
0 & \text{otherwise}.
\end{cases}
\end{equation}

\hspace{1.0em} \textit{b) Strict Prefix Token Matching}

Let ($\mathbf{y} = \text{tok}(m)$) be the tokenized ground-truth span, ($\hat{\mathbf{y}} = \text{tok}(o_{\text{answer}})$) be the tokenized answer. We define \textbf{strict prefix match} as:
\begin{equation}
    \text{PrefixMatch}(\hat{\mathbf{y}}, \mathbf{y}) =
\begin{cases}
1 & \text{if } \hat{\mathbf{y}} \text{ is a strict prefix of } \mathbf{y}, \\
0 & \text{otherwise}.
\end{cases}
\end{equation}

This design serves two important purposes: 1) It avoids penalizing the model when it predicts an \textbf{incomplete but correct beginning} of the masked span.
2) It reduces brittleness caused by minor tokenizer-level variations, while still preventing over-generous scoring for semantically unrelated outputs.

\hspace{1.0em} \textit{c) Final Answer Reward}

The correctness score is then:
\begin{equation}
    r_{\text{ans}} =
\max(\text{EM}, \text{PrefixMatch}),
\end{equation}

ensuring that the model receives reward as long as it produces a \textbf{correct prefix} of the masked content-even if the full span is not reproduced verbatim.

This softer yet principled correctness metric is particularly useful when the masked unit is long or contains tokens whose boundaries are tokenizer-dependent.

\section{Experiment}
\subsection{Experiment Setup}
\textbf{Masked-pretraining data}: We use the publicly available image–caption set \textbf{ALLaVA-Caption-LAION-4V}\cite{chen2024allava} as the source corpus for mask construction, which provides sufficiently diverse visual contexts and rich natural language descriptions.

\textbf{Proxy Multimodal Model for Visual Attention Measurement}: To compute sentence-level visual dependency scores (Section~\ref{dentifying Vision-Dependent Language Units}), we adopt Qwen2.5-VL-7B \cite{bai2025qwen2} as the proxy vision–language model. Following analyses in prior work on attention grounding and multimodal interpretability\cite{kang2025your,esmaeilkhani2025direct,yang2025learning}, we use several last few layers of all the decoder layers (\textit{e.g.}, the last 3 to 6 of all the 36 layers) for visual dependency computation. These bottom layers have been shown to best capture \textit{semantic-level cross-modal interactions}, making them appropriate for identifying genuinely vision-critical language spans.

\textbf{Text-Only LLM for Fine-Grained Visual-Need Annotation}: To refine visually dependent units within each sentence (Section~\ref{dentifying Vision-Dependent Language Units}), we employ \textbf{Qwen2.5-Instruct-72B} \cite{bai2025qwen2}, a high-capacity text-only LLM with strong semantic reasoning abilities. The model receives the raw caption text and outputs a version annotated with \{...\} brackets marking tokens or phrases that cannot be reliably inferred without visual information.

\textbf{RL Framework}: Our reinforcement learning pipeline is built on Easy-R1\cite{zheng2025easyr1, sheng2024hybridflow}, a training framework designed for scalable reasoning-oriented RL, combined with Verl\cite{sheng2024hybridflow}, a modular and production-grade RL engine for large language models.

\textbf{Benchmarks}: We evaluate models, using VLMEvalKit \cite{duan2025vlmevalkitopensourcetoolkitevaluating}, on a diverse suite of seven multimodal reasoning benchmarks spanning perception, visual grounding, language–vision alignment, and numerical or symbolic inference. 
\begin{itemize}
    \item \textbf{MMVet} \cite{yu2024mmvetevaluatinglargemultimodal}, \textbf{MMBench-v1.1} \cite{liu2024mmbenchmultimodalmodelallaround}, and \textbf{MMStar} \cite{chen2024rightwayevaluatinglarge} measure general-purpose visual understanding and multi-step reasoning across real-world images;
    \item \textbf{BLINK} \cite{fu2024blinkmultimodallargelanguage} evaluates fine-grained visual grounding and disambiguation under minimal textual priors.;
    \item \textbf{MathVista-MINI} \cite{lu2024mathvistaevaluatingmathematicalreasoning}  and \textbf{WeMath} \cite{qiao2024wemathdoeslargemultimodal} focus on visual–mathematical reasoning involving diagrams, equations, and implicit numeric cues;
    \item \textbf{ChartQA} \cite{masry2022chartqabenchmarkquestionanswering} examines structured comprehension of charts and plots, emphasizing cross-modal parsing and factual consistency.
\end{itemize}
Together, these benchmarks cover a broad spectrum of multimodal competencies and provide a comprehensive assessment of visual grounding, cross-modal reasoning, and robustness.

\subsection{Layer-wise Analysis of Visual Dependency}
To determine which decoder layers provide the most reliable signal for visual-dependency estimation, we probe Qwen2.5-VL-7B on 5k captions and compute token-level visual attention ratios for every layer. For each caption, we segment the text into sentences and derive two simple statistics at layer ($l$):
\begin{itemize}
    \item \textbf{Within-Sentence Variance ($\sigma_{\text{within}}(l)$)}: the average token-level variance inside each sentence, measuring how strongly a layer highlights \textbf{localized visual-dependent spans} within a sentence;
    \item \textbf{Between-Sentence Variance ($\sigma_{\text{between}}(l)$}: the variance of sentence-level means within the same caption, indicating how well a layer separates \textbf{vision-driven sentences} from \textbf{language-driven ones}.
\end{itemize}
The ideal layers for estimating visual dependency should (i) show \textbf{clear token-level contrast} inside truly vision-dependent sentences, and (ii) yield \textbf{large differences} across sentences of a caption, since mixed captions typically contain both visually grounded and purely linguistic segments.

\begin{figure}
    \centering
    \includegraphics[width=0.86\linewidth]{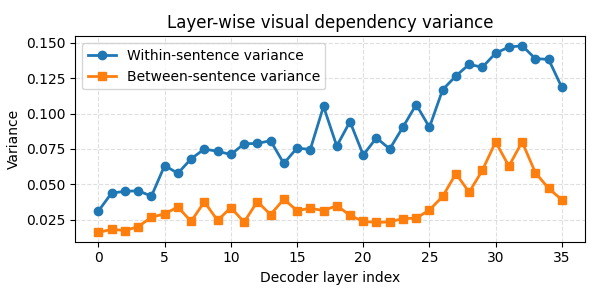}
    \caption{Layer-wise within- and between-sentence variance of token-level visual dependency in Qwen2.5-VL-7B. Both metrics peak in the upper-middle decoder layers (29–31), indicating that these layers best differentiate visually grounded content from language-only content. We therefore adopt this layer region for dependency estimation in MMRPT.}
    \label{fig:layer-attn}
\end{figure}

Figure~\ref{fig:layer-attn} reports ($\sigma_{\text{within}}(l)$) and ($\sigma_{\text{between}}(l)$) across all decoder layers.
Both within-sentence and between-sentence variance exhibit a characteristic “semantic-alignment pattern”: early layers remain low, mid layers gradually increase, and upper-middle layers (roughly layer 29–31) reach the highest contrast. This matches the common observation that cross-modal alignment emerges in the upper-middle decoder layers of VL models, whereas the final layers increasingly favor linguistic decoding. Qualitative inspection further confirms that these high-contrast spans correspond to salient objects, on-image text, or simple visual attributes. Based on this trend, we select the last 4–6 layers for visual-dependency estimation.

\subsection{Main Results}
To assess the effect of MMRPT as a \textit{pre-training paradigm} rather than as a tuning trick, we directly compare the original model (\textbf{Base}) and the model after masked multimodal reinforcement pre-training (\textbf{MMRPT}) under identical conditions. Neither model receives supervised fine-tuning in this comparison. Both models are evaluated in the zero-shot setting. 

Results in Table~\ref{tab:zeroshot_multi_model} show that MMRPT delivers consistent zero-shot gains over the Base model across nearly all benchmarks, with absolute improvements typically in the 0.5–1\% range. While the magnitude is modest, the uniformity of the gains indicates that masked multimodal reinforcement pre-training strengthens the model’s ability to extract and use visual evidence without additional supervision. The only task showing no clear benefit is ChartQA, whose highly structured chart-parsing nature relies less on the visual grounding mechanisms targeted by MMRPT. Overall, these results demonstrate that MMRPT provides a stable and generalizable enhancement to multimodal reasoning.

\begin{table*}[t]
\centering
\small
\setlength{\tabcolsep}{4.5pt}
\begin{tabular}{lccccccc}
\toprule
& \textbf{MMVet} & \textbf{MMBench} & \textbf{MMStar} &
\textbf{BLINK} & \textbf{MathVista} & \textbf{WeMath} &
\textbf{ChartQA} \\
\midrule
\multicolumn{8}{c}{\textbf{Qwen2.5-VL-3B-Instruct}} \\
\midrule
Base   &0.5940\color{white}{↑} &0.7996\color{white}{↑} &0.5604\color{white}{↑} &0.4890\color{white}{↑} &0.6230\color{white}{↑} &0.2381\color{white}{↑} &0.8408\color{white}{↑}    \\
MMRPT  & {\color{green!60!black}{0.6062↑}} & {\color{green!60!black}{0.8056↑}} &
         {\color{green!60!black}{0.5819↑}} & {\color{green!60!black}{0.4940↑}} &
         {\color{green!60!black}{0.6350↑}} & {\color{green!60!black}{0.2464↑}} &
         {\color{red!60!black}{0.8396↓}} \\
\midrule
\multicolumn{8}{c}{\textbf{Qwen2.5-VL-7B-Instruct}} \\
\midrule
Base   &    0.6740&    0.8259&    0.6390&    0.5513&    0.6820&    0.3596&    0.8725\\
MMRPT  & {\color{green!60!black}{0.6812↑ }} & {\color{green!60!black}{0.8336↑ }} &
         {\color{green!60!black}{0.6524↑ }} & {\color{green!60!black}{0.5589↑ }} &
         {\color{green!60!black}{0.6904↑ }} & {\color{green!60!black}{0.3688↑ }} &
         {\color{green!60!black}{0.8733↑ }} \\
\bottomrule
\end{tabular}
\caption{
Zero-shot comparison between the original Base model and the MMRPT-enhanced model across multiple Qwen2.5-VL scales. Green (↑) indicates performance improvement; red (↓) indicates a decrease.}
\label{tab:zeroshot_multi_model}
\end{table*}

To assess whether MMRPT provides a better initialization for downstream learning, we fine-tune both the original model (Base+SFT) and the MMRPT-pretrained model (MMRPT+SFT) on a 100k curated WeMath-style corpus distilled from the WeThink \cite{yang2025wethinkgeneralpurposevisionlanguagereasoning} dataset using GLM-4.5V \cite{vteam2025glm45vglm41vthinkingversatilemultimodal} and Seed-1.5VL \cite{guo2025seed15vltechnicalreport}, ensuring high-quality multi-step visual–math reasoning traces under identical optimization settings.
Both variants use the same architecture, tokenizer, multimodal adapters, training corpus, optimization schedule, and hyperparameters; the only difference lies in whether MMRPT is applied before fine-tuning. We evaluate all models again, as Table~\ref{tab:sft_mmrpt} shows, on the same benchmark suite as in the zero-shot setting.

\begin{table*}[t]
\centering
\small
\setlength{\tabcolsep}{4.5pt}
\begin{tabular}{lccccccc}
\toprule
 & \textbf{MMVet} & \textbf{MMBench} & \textbf{MMStar} &
   \textbf{BLINK} & \textbf{MathVista} & \textbf{WeMath} &
   \textbf{ChartQA} \\
\midrule
\multicolumn{8}{c}{\textbf{Qwen2.5-VL-3B-Instruct}} \\
\midrule
Base+SFT      &   \color{green!60!black}{0.6422↑}&   \color{red!60!black}{0.7908↓}&  \color{green!60!black}{0.5834↑}&    \color{green!60!black}{0.5016↑}&   \color{red!60!black}{0.5920↓}&   \color{green!60!black}{0.2502↑}&   \color{red!60!black}{0.6769↓}\\
MMRPT+SFT     &   \color{green!60!black}{0.6508↑ }&   \color{green!60!black}{0.8074↑}&   \color{green!60!black}{0.6012↑}&   \color{green!60!black}{0.5114↑}&   \color{green!60!black}{0.6406↑}&   \color{green!60!black}{0.2614↑}&   \color{green!60!black}{0.8366↑}\\
\midrule
\multicolumn{8}{c}{\textbf{Qwen2.5-VL-7B-Instruct}} \\
\midrule  
Base+SFT      &   \color{green!60!black}{0.7266↑}&   \color{red!60!black}{0.8150↓}&   \color{green!60!black}{0.6477↑}&   \color{green!60!black}{0.5724↑}&   \color{red!60!black}{0.6612↓}&   \color{green!60!black}{0.3962↑}&  \color{red!60!black}{0.8125↓} \\
MMRPT+SFT     &   \color{green!60!black}{0.7398↑}&   \color{green!60!black}{0.8296↑}&   \color{green!60!black}{0.6561↑}&   \color{green!60!black}{0.5788↑}&   \color{green!60!black}{0.7023↑}&   \color{green!60!black}{0.4006↑}&   \color{green!60!black}{0.8798↑}\\
\bottomrule
\end{tabular}
\caption{
Fine-tuned performance comparison between the original Base models (\textbf{Base+SFT})
and the MMRPT-enhanced models (\textbf{MMRPT+SFT}) under identical supervised
training conditions across multiple Qwen2.5-VL scales.
All models are evaluated on the same set of multimodal reasoning benchmarks.}
\label{tab:sft_mmrpt}
\end{table*}

After supervised fine-tuning on domain-specific data (e.g., WeMath), the Base model exhibits substantial in-domain gains but suffers pronounced degradation on out-of-distribution benchmarks such as MMBench, MathVista, and ChartQA, indicating a clear overfitting tendency and a loss of general multimodal competence. In contrast, the MMRPT-pretrained model maintains stable performance across all OOD evaluations under the same fine-tuning setup, while still achieving comparable or slightly higher in-domain improvements. This demonstrates that MMRPT provides a more robust initialization for downstream learning, effectively preventing catastrophic degradation and preserving general reasoning ability even after task-specialized fine-tuning.
\section{Conclusion}
We introduced MMRPT, a masked multimodal reinforcement pre-training framework designed to strengthen visual grounding and cross-modal reasoning in large vision–language models. By identifying vision-dependent spans and training the model to reconstruct them through RL-guided reasoning, MMRPT pushes the model beyond caption imitation toward more faithful visual understanding. Experiments across diverse multimodal benchmarks show consistent zero-shot gains and substantially improved robustness under supervised fine-tuning, where MMRPT prevents the degradation commonly observed in standard models. These results demonstrate that integrating reinforcement-driven masked reasoning into the pre-training pipeline offers a simple yet effective path toward more reliable and generalizable multimodal intelligence.

% \section*{References}
\bibliographystyle{unsrt}
\bibliography{ref.bib}

@article{li2021align,
  title={Align before fuse: Vision and language representation learning with momentum distillation},
  author={Li, Junnan and Selvaraju, Ramprasaath and Gotmare, Akhilesh and Joty, Shafiq and Xiong, Caiming and Hoi, Steven Chu Hong},
  journal={Advances in neural information processing systems},
  volume={34},
  pages={9694--9705},
  year={2021}
}

@inproceedings{lewis2024does,
  title={Does clip bind concepts? probing compositionality in large image models},
  author={Lewis, Martha and Nayak, Nihal and Yu, Peilin and Merullo, Jack and Yu, Qinan and Bach, Stephen and Pavlick, Ellie},
  booktitle={Findings of the Association for Computational Linguistics: EACL 2024},
  pages={1487--1500},
  year={2024}
}

@misc{liu2023visualinstructiontuning,
      title={Visual Instruction Tuning}, 
      author={Haotian Liu and Chunyuan Li and Qingyang Wu and Yong Jae Lee},
      year={2023},
      eprint={2304.08485},
      archivePrefix={arXiv},
      primaryClass={cs.CV},
      url={https://arxiv.org/abs/2304.08485}, 
}

@inproceedings{radford2021learning,
  title={Learning transferable visual models from natural language supervision},
  author={Radford, Alec and Kim, Jong Wook and Hallacy, Chris and Ramesh, Aditya and Goh, Gabriel and Agarwal, Sandhini and Sastry, Girish and Askell, Amanda and Mishkin, Pamela and Clark, Jack and others},
  booktitle={International conference on machine learning},
  pages={8748--8763},
  year={2021},
  organization={PmLR}
}

@article{yin2024survey,
  title={A survey on multimodal large language models},
  author={Yin, Shukang and Fu, Chaoyou and Zhao, Sirui and Li, Ke and Sun, Xing and Xu, Tong and Chen, Enhong},
  journal={National Science Review},
  volume={11},
  number={12},
  pages={nwae403},
  year={2024},
  publisher={Oxford University Press}
}

@inproceedings{chen2024lion,
  title={Lion: Empowering multimodal large language model with dual-level visual knowledge},
  author={Chen, Gongwei and Shen, Leyang and Shao, Rui and Deng, Xiang and Nie, Liqiang},
  booktitle={Proceedings of the IEEE/CVF Conference on Computer Vision and Pattern Recognition},
  pages={26540--26550},
  year={2024}
}

@article{ghosh2024exploring,
  title={Exploring the frontier of vision-language models: A survey of current methodologies and future directions},
  author={Ghosh, Akash and Acharya, Arkadeep and Saha, Sriparna and Jain, Vinija and Chadha, Aman},
  journal={arXiv preprint arXiv:2404.07214},
  year={2024}
}

@article{jin2024efficient,
  title={Efficient multimodal large language models: A survey},
  author={Jin, Yizhang and Li, Jian and Liu, Yexin and Gu, Tianjun and Wu, Kai and Jiang, Zhengkai and He, Muyang and Zhao, Bo and Tan, Xin and Gan, Zhenye and others},
  journal={arXiv preprint arXiv:2405.10739},
  year={2024}
}

@article{caffagni2024revolution,
  title={The revolution of multimodal large language models: a survey},
  author={Caffagni, Davide and Cocchi, Federico and Barsellotti, Luca and Moratelli, Nicholas and Sarto, Sara and Baraldi, Lorenzo and Cornia, Marcella and Cucchiara, Rita},
  journal={arXiv preprint arXiv:2402.12451},
  year={2024}
}

@article{schuhmann2022laion,
  title={Laion-5b: An open large-scale dataset for training next generation image-text models},
  author={Schuhmann, Christoph and Beaumont, Romain and Vencu, Richard and Gordon, Cade and Wightman, Ross and Cherti, Mehdi and Coombes, Theo and Katta, Aarush and Mullis, Clayton and Wortsman, Mitchell and others},
  journal={Advances in neural information processing systems},
  volume={35},
  pages={25278--25294},
  year={2022}
}

@article{yu2022coca,
  title={Coca: Contrastive captioners are image-text foundation models},
  author={Yu, Jiahui and Wang, Zirui and Vasudevan, Vijay and Yeung, Legg and Seyedhosseini, Mojtaba and Wu, Yonghui},
  journal={arXiv preprint arXiv:2205.01917},
  year={2022}
}

@inproceedings{chen2024internvl,
  title={Internvl: Scaling up vision foundation models and aligning for generic visual-linguistic tasks},
  author={Chen, Zhe and Wu, Jiannan and Wang, Wenhai and Su, Weijie and Chen, Guo and Xing, Sen and Zhong, Muyan and Zhang, Qinglong and Zhu, Xizhou and Lu, Lewei and others},
  booktitle={Proceedings of the IEEE/CVF conference on computer vision and pattern recognition},
  pages={24185--24198},
  year={2024}
}

@article{bai2025qwen2,
  title={Qwen2. 5-vl technical report},
  author={Bai, Shuai and Chen, Keqin and Liu, Xuejing and Wang, Jialin and Ge, Wenbin and Song, Sibo and Dang, Kai and Wang, Peng and Wang, Shijie and Tang, Jun and others},
  journal={arXiv preprint arXiv:2502.13923},
  year={2025}
}

@misc{vteam2025glm45vglm41vthinkingversatilemultimodal,
      title={GLM-4.5V and GLM-4.1V-Thinking: Towards Versatile Multimodal Reasoning with Scalable Reinforcement Learning}, 
      author={V Team and Wenyi Hong and Wenmeng Yu and Xiaotao Gu and Guo Wang and Guobing Gan and Haomiao Tang and Jiale Cheng and Ji Qi and Junhui Ji and Lihang Pan and Shuaiqi Duan and Weihan Wang and Yan Wang and Yean Cheng and Zehai He and Zhe Su and Zhen Yang and Ziyang Pan and Aohan Zeng and Baoxu Wang and Bin Chen and Boyan Shi and Changyu Pang and Chenhui Zhang and Da Yin and Fan Yang and Guoqing Chen and Jiazheng Xu and Jiale Zhu and Jiali Chen and Jing Chen and Jinhao Chen and Jinghao Lin and Jinjiang Wang and Junjie Chen and Leqi Lei and Letian Gong and Leyi Pan and Mingdao Liu and Mingde Xu and Mingzhi Zhang and Qinkai Zheng and Sheng Yang and Shi Zhong and Shiyu Huang and Shuyuan Zhao and Siyan Xue and Shangqin Tu and Shengbiao Meng and Tianshu Zhang and Tianwei Luo and Tianxiang Hao and Tianyu Tong and Wenkai Li and Wei Jia and Xiao Liu and Xiaohan Zhang and Xin Lyu and Xinyue Fan and Xuancheng Huang and Yanling Wang and Yadong Xue and Yanfeng Wang and Yanzi Wang and Yifan An and Yifan Du and Yiming Shi and Yiheng Huang and Yilin Niu and Yuan Wang and Yuanchang Yue and Yuchen Li and Yutao Zhang and Yuting Wang and Yu Wang and Yuxuan Zhang and Zhao Xue and Zhenyu Hou and Zhengxiao Du and Zihan Wang and Peng Zhang and Debing Liu and Bin Xu and Juanzi Li and Minlie Huang and Yuxiao Dong and Jie Tang},
      year={2025},
      eprint={2507.01006},
      archivePrefix={arXiv},
      primaryClass={cs.CV},
      url={https://arxiv.org/abs/2507.01006}, 
}

@misc{yu2025minicpmv45cookingefficient,
      title={MiniCPM-V 4.5: Cooking Efficient MLLMs via Architecture, Data, and Training Recipe}, 
      author={Tianyu Yu and Zefan Wang and Chongyi Wang and Fuwei Huang and Wenshuo Ma and Zhihui He and Tianchi Cai and Weize Chen and Yuxiang Huang and Yuanqian Zhao and Bokai Xu and Junbo Cui and Yingjing Xu and Liqing Ruan and Luoyuan Zhang and Hanyu Liu and Jingkun Tang and Hongyuan Liu and Qining Guo and Wenhao Hu and Bingxiang He and Jie Zhou and Jie Cai and Ji Qi and Zonghao Guo and Chi Chen and Guoyang Zeng and Yuxuan Li and Ganqu Cui and Ning Ding and Xu Han and Yuan Yao and Zhiyuan Liu and Maosong Sun},
      year={2025},
      eprint={2509.18154},
      archivePrefix={arXiv},
      primaryClass={cs.LG},
      url={https://arxiv.org/abs/2509.18154}, 
}

@misc{coreteam2025mimovltechnicalreport,
      title={MiMo-VL Technical Report}, 
      author={Core Team and Zihao Yue and Zhenru Lin and Yifan Song and Weikun Wang and Shuhuai Ren and Shuhao Gu and Shicheng Li and Peidian Li and Liang Zhao and Lei Li and Kainan Bao and Hao Tian and Hailin Zhang and Gang Wang and Dawei Zhu and Cici and Chenhong He and Bowen Ye and Bowen Shen and Zihan Zhang and Zihan Jiang and Zhixian Zheng and Zhichao Song and Zhenbo Luo and Yue Yu and Yudong Wang and Yuanyuan Tian and Yu Tu and Yihan Yan and Yi Huang and Xu Wang and Xinzhe Xu and Xingchen Song and Xing Zhang and Xing Yong and Xin Zhang and Xiangwei Deng and Wenyu Yang and Wenhan Ma and Weiwei Lv and Weiji Zhuang and Wei Liu and Sirui Deng and Shuo Liu and Shimao Chen and Shihua Yu and Shaohui Liu and Shande Wang and Rui Ma and Qiantong Wang and Peng Wang and Nuo Chen and Menghang Zhu and Kangyang Zhou and Kang Zhou and Kai Fang and Jun Shi and Jinhao Dong and Jiebao Xiao and Jiaming Xu and Huaqiu Liu and Hongshen Xu and Heng Qu and Haochen Zhao and Hanglong Lv and Guoan Wang and Duo Zhang and Dong Zhang and Di Zhang and Chong Ma and Chang Liu and Can Cai and Bingquan Xia},
      year={2025},
      eprint={2506.03569},
      archivePrefix={arXiv},
      primaryClass={cs.CL},
      url={https://arxiv.org/abs/2506.03569}, 
}

@inproceedings{sim2025can,
  title={Can vlms actually see and read? a survey on modality collapse in vision-language models},
  author={Sim, Mong Yuan and Zhang, Wei Emma and Dai, Xiang and Fang, Biaoyan},
  booktitle={Findings of the Association for Computational Linguistics: ACL 2025},
  pages={24452--24470},
  year={2025}
}

@article{zhang2024visually,
  title={Why are visually-grounded language models bad at image classification?},
  author={Zhang, Yuhui and Unell, Alyssa and Wang, Xiaohan and Ghosh, Dhruba and Su, Yuchang and Schmidt, Ludwig and Yeung-Levy, Serena},
  journal={Advances in Neural Information Processing Systems},
  volume={37},
  pages={51727--51753},
  year={2024}
}

@article{vo2025vision,
  title={Vision Language Models are Biased},
  author={Vo, An and Nguyen, Khai-Nguyen and Taesiri, Mohammad Reza and Dang, Vy Tuong and Nguyen, Anh Totti and Kim, Daeyoung},
  journal={arXiv preprint arXiv:2505.23941},
  year={2025}
}

@article{bai2024hallucination,
  title={Hallucination of multimodal large language models: A survey},
  author={Bai, Zechen and Wang, Pichao and Xiao, Tianjun and He, Tong and Han, Zongbo and Zhang, Zheng and Shou, Mike Zheng},
  journal={arXiv preprint arXiv:2404.18930},
  year={2024}
}

@article{chen2024multi,
  title={Multi-object hallucination in vision language models},
  author={Chen, Xuweiyi and Ma, Ziqiao and Zhang, Xuejun and Xu, Sihan and Qian, Shengyi and Yang, Jianing and Fouhey, David and Chai, Joyce},
  journal={Advances in Neural Information Processing Systems},
  volume={37},
  pages={44393--44418},
  year={2024}
}

@inproceedings{zhao2025looking,
  title={Looking beyond text: Reducing language bias in large vision-language models via multimodal dual-attention and soft-image guidance},
  author={Zhao, Haozhe and Si, Shuzheng and Chen, Liang and Zhang, Yichi and Sun, Maosong and Chang, Baobao and Zhang, Minjia},
  booktitle={Proceedings of the 2025 Conference on Empirical Methods in Natural Language Processing},
  pages={19677--19701},
  year={2025}
}

@article{park2025second,
  title={SECOND: Mitigating Perceptual Hallucination in Vision-Language Models via Selective and Contrastive Decoding},
  author={Park, Woohyeon and Kim, Woojin and Kim, Jaeik and Do, Jaeyoung},
  journal={arXiv preprint arXiv:2506.08391},
  year={2025}
}

@article{lin2023revisiting,
  title={Revisiting the role of language priors in vision-language models},
  author={Lin, Zhiqiu and Chen, Xinyue and Pathak, Deepak and Zhang, Pengchuan and Ramanan, Deva},
  journal={arXiv preprint arXiv:2306.01879},
  year={2023}
}

@article{zhao2025mitigate,
  title={Mitigate Language Priors in Large Vision-Language Models by Cross-Images Contrastive Decoding},
  author={Zhao, Jianfei and Zhang, Feng and Sun, Xin and Feng, Chong},
  journal={arXiv e-prints},
  pages={arXiv--2505},
  year={2025}
}

@article{zheng2025modality,
  title={Modality Bias in LVLMs: Analyzing and Mitigating Object Hallucination via Attention Lens},
  author={Zheng, Haohan and Zhang, Zhenguo},
  journal={arXiv preprint arXiv:2508.02419},
  year={2025}
}

@inproceedings{dai2025see,
  title={See Different, Think Better: Visual Variations Mitigating Hallucinations in LVLMs},
  author={Dai, Ziyun and Li, Xiaoqiang and Zhang, Shaohua and Wu, Yuanchen and Li, Jide},
  booktitle={Proceedings of the 33rd ACM International Conference on Multimedia},
  pages={3310--3319},
  year={2025}
}

@article{zhou2025perception,
  title={From perception to cognition: A survey of vision-language interactive reasoning in multimodal large language models},
  author={Zhou, Chenyue and Wang, Mingxuan and Ma, Yanbiao and Wu, Chenxu and Chen, Wanyi and Qian, Zhe and Liu, Xinyu and Zhang, Yiwei and Wang, Junhao and Xu, Hengbo and others},
  journal={arXiv preprint arXiv:2509.25373},
  year={2025}
}

@article{islam2024large,
  title={Are large vision language models up to the challenge of chart comprehension and reasoning? an extensive investigation into the capabilities and limitations of lvlms},
  author={Islam, Mohammed Saidul and Rahman, Raian and Masry, Ahmed and Laskar, Md Tahmid Rahman and Nayeem, Mir Tafseer and Hoque, Enamul},
  journal={arXiv preprint arXiv:2406.00257},
  year={2024}
}

@article{zheng2024thinking,
  title={Thinking before looking: Improving multimodal llm reasoning via mitigating visual hallucination},
  author={Zheng, Haojie and Xu, Tianyang and Sun, Hanchi and Pu, Shu and Chen, Ruoxi and Sun, Lichao},
  journal={arXiv preprint arXiv:2411.12591},
  year={2024}
}

@inproceedings{kang2025your,
  title={Your large vision-language model only needs a few attention heads for visual grounding},
  author={Kang, Seil and Kim, Jinyeong and Kim, Junhyeok and Hwang, Seong Jae},
  booktitle={Proceedings of the Computer Vision and Pattern Recognition Conference},
  pages={9339--9350},
  year={2025}
}

@article{esmaeilkhani2025direct,
  title={Direct Visual Grounding by Directing Attention of Visual Tokens},
  author={Esmaeilkhani, Parsa and Latecki, Longin Jan},
  journal={arXiv preprint arXiv:2511.12738},
  year={2025}
}

@article{yang2025learning,
  title={Learning to Look: Cognitive Attention Alignment with Vision-Language Models},
  author={Yang, Ryan L and Bhusal, Dipkamal and Rastogi, Nidhi},
  journal={arXiv preprint arXiv:2509.21247},
  year={2025}
}

@misc{zheng2025easyr1,
  title        = {EasyR1: An Efficient, Scalable, Multi-Modality RL Training Framework},
  author       = {Yaowei Zheng, Junting Lu and Shenzhi Wang and Zhangchi Feng and Dongdong Kuang and Yuwen Xiong},
  howpublished = {\url{https://github.com/hiyouga/EasyR1}},
  year         = {2025}
}

@article{sheng2024hybridflow,
  title   = {HybridFlow: A Flexible and Efficient RLHF Framework},
  author  = {Guangming Sheng and Chi Zhang and Zilingfeng Ye and Xibin Wu and Wang Zhang and Ru Zhang and Yanghua Peng and Haibin Lin and Chuan Wu},
  year    = {2024},
  journal = {arXiv preprint arXiv: 2409.19256}
}

@misc{chen2024allava,
      title={ALLaVA: Harnessing GPT4V-synthesized Data for A Lite Vision-Language Model}, 
      author={Guiming Hardy Chen and Shunian Chen and Ruifei Zhang and Junying Chen and Xiangbo Wu and Zhiyi Zhang and Zhihong Chen and Jianquan Li and Xiang Wan and Benyou Wang},
      year={2024},
      eprint={2402.11684},
      archivePrefix={arXiv},
      primaryClass={cs.CL}
}

@misc{yu2024mmvetevaluatinglargemultimodal,
      title={MM-Vet: Evaluating Large Multimodal Models for Integrated Capabilities}, 
      author={Weihao Yu and Zhengyuan Yang and Linjie Li and Jianfeng Wang and Kevin Lin and Zicheng Liu and Xinchao Wang and Lijuan Wang},
      year={2024},
      eprint={2308.02490},
      archivePrefix={arXiv},
      primaryClass={cs.AI},
      url={https://arxiv.org/abs/2308.02490}, 
}

@misc{liu2024mmbenchmultimodalmodelallaround,
      title={MMBench: Is Your Multi-modal Model an All-around Player?}, 
      author={Yuan Liu and Haodong Duan and Yuanhan Zhang and Bo Li and Songyang Zhang and Wangbo Zhao and Yike Yuan and Jiaqi Wang and Conghui He and Ziwei Liu and Kai Chen and Dahua Lin},
      year={2024},
      eprint={2307.06281},
      archivePrefix={arXiv},
      primaryClass={cs.CV},
      url={https://arxiv.org/abs/2307.06281}, 
}

@misc{chen2024rightwayevaluatinglarge,
      title={Are We on the Right Way for Evaluating Large Vision-Language Models?}, 
      author={Lin Chen and Jinsong Li and Xiaoyi Dong and Pan Zhang and Yuhang Zang and Zehui Chen and Haodong Duan and Jiaqi Wang and Yu Qiao and Dahua Lin and Feng Zhao},
      year={2024},
      eprint={2403.20330},
      archivePrefix={arXiv},
      primaryClass={cs.CV},
      url={https://arxiv.org/abs/2403.20330}, 
}

@misc{fu2024blinkmultimodallargelanguage,
      title={BLINK: Multimodal Large Language Models Can See but Not Perceive}, 
      author={Xingyu Fu and Yushi Hu and Bangzheng Li and Yu Feng and Haoyu Wang and Xudong Lin and Dan Roth and Noah A. Smith and Wei-Chiu Ma and Ranjay Krishna},
      year={2024},
      eprint={2404.12390},
      archivePrefix={arXiv},
      primaryClass={cs.CV},
      url={https://arxiv.org/abs/2404.12390}, 
}

@misc{lu2024mathvistaevaluatingmathematicalreasoning,
      title={MathVista: Evaluating Mathematical Reasoning of Foundation Models in Visual Contexts}, 
      author={Pan Lu and Hritik Bansal and Tony Xia and Jiacheng Liu and Chunyuan Li and Hannaneh Hajishirzi and Hao Cheng and Kai-Wei Chang and Michel Galley and Jianfeng Gao},
      year={2024},
      eprint={2310.02255},
      archivePrefix={arXiv},
      primaryClass={cs.CV},
      url={https://arxiv.org/abs/2310.02255}, 
}

@misc{qiao2024wemathdoeslargemultimodal,
      title={We-Math: Does Your Large Multimodal Model Achieve Human-like Mathematical Reasoning?}, 
      author={Runqi Qiao and Qiuna Tan and Guanting Dong and Minhui Wu and Chong Sun and Xiaoshuai Song and Zhuoma GongQue and Shanglin Lei and Zhe Wei and Miaoxuan Zhang and Runfeng Qiao and Yifan Zhang and Xiao Zong and Yida Xu and Muxi Diao and Zhimin Bao and Chen Li and Honggang Zhang},
      year={2024},
      eprint={2407.01284},
      archivePrefix={arXiv},
      primaryClass={cs.AI},
      url={https://arxiv.org/abs/2407.01284}, 
}

@misc{masry2022chartqabenchmarkquestionanswering,
      title={ChartQA: A Benchmark for Question Answering about Charts with Visual and Logical Reasoning}, 
      author={Ahmed Masry and Do Xuan Long and Jia Qing Tan and Shafiq Joty and Enamul Hoque},
      year={2022},
      eprint={2203.10244},
      archivePrefix={arXiv},
      primaryClass={cs.CL},
      url={https://arxiv.org/abs/2203.10244}, 
}

@misc{yang2025wethinkgeneralpurposevisionlanguagereasoning,
      title={WeThink: Toward General-purpose Vision-Language Reasoning via Reinforcement Learning}, 
      author={Jie Yang and Feipeng Ma and Zitian Wang and Dacheng Yin and Kang Rong and Fengyun Rao and Ruimao Zhang},
      year={2025},
      eprint={2506.07905},
      archivePrefix={arXiv},
      primaryClass={cs.CV},
      url={https://arxiv.org/abs/2506.07905}, 
}

@misc{guo2025seed15vltechnicalreport,
      title={Seed1.5-VL Technical Report}, 
      author={Dong Guo and Faming Wu and Feida Zhu and Fuxing Leng and Guang Shi and Haobin Chen and Haoqi Fan and Jian Wang and Jianyu Jiang and Jiawei Wang and Jingji Chen and Jingjia Huang and Kang Lei and Liping Yuan and Lishu Luo and Pengfei Liu and Qinghao Ye and Rui Qian and Shen Yan and Shixiong Zhao and Shuai Peng and Shuangye Li and Sihang Yuan and Sijin Wu and Tianheng Cheng and Weiwei Liu and Wenqian Wang and Xianhan Zeng and Xiao Liu and Xiaobo Qin and Xiaohan Ding and Xiaojun Xiao and Xiaoying Zhang and Xuanwei Zhang and Xuehan Xiong and Yanghua Peng and Yangrui Chen and Yanwei Li and Yanxu Hu and Yi Lin and Yiyuan Hu and Yiyuan Zhang and Youbin Wu and Yu Li and Yudong Liu and Yue Ling and Yujia Qin and Zanbo Wang and Zhiwu He and Aoxue Zhang and Bairen Yi and Bencheng Liao and Can Huang and Can Zhang and Chaorui Deng and Chaoyi Deng and Cheng Lin and Cheng Yuan and Chenggang Li and Chenhui Gou and Chenwei Lou and Chengzhi Wei and Chundian Liu and Chunyuan Li and Deyao Zhu and Donghong Zhong and Feng Li and Feng Zhang and Gang Wu and Guodong Li and Guohong Xiao and Haibin Lin and Haihua Yang and Haoming Wang and Heng Ji and Hongxiang Hao and Hui Shen and Huixia Li and Jiahao Li and Jialong Wu and Jianhua Zhu and Jianpeng Jiao and Jiashi Feng and Jiaze Chen and Jianhui Duan and Jihao Liu and Jin Zeng and Jingqun Tang and Jingyu Sun and Joya Chen and Jun Long and Junda Feng and Junfeng Zhan and Junjie Fang and Junting Lu and Kai Hua and Kai Liu and Kai Shen and Kaiyuan Zhang and Ke Shen and Ke Wang and Keyu Pan and Kun Zhang and Kunchang Li and Lanxin Li and Lei Li and Lei Shi and Li Han and Liang Xiang and Liangqiang Chen and Lin Chen and Lin Li and Lin Yan and Liying Chi and Longxiang Liu and Mengfei Du and Mingxuan Wang and Ningxin Pan and Peibin Chen and Pengfei Chen and Pengfei Wu and Qingqing Yuan and Qingyao Shuai and Qiuyan Tao and Renjie Zheng and Renrui Zhang and Ru Zhang and Rui Wang and Rui Yang and Rui Zhao and Shaoqiang Xu and Shihao Liang and Shipeng Yan and Shu Zhong and Shuaishuai Cao and Shuangzhi Wu and Shufan Liu and Shuhan Chang and Songhua Cai and Tenglong Ao and Tianhao Yang and Tingting Zhang and Wanjun Zhong and Wei Jia and Wei Weng and Weihao Yu and Wenhao Huang and Wenjia Zhu and Wenli Yang and Wenzhi Wang and Xiang Long and XiangRui Yin and Xiao Li and Xiaolei Zhu and Xiaoying Jia and Xijin Zhang and Xin Liu and Xinchen Zhang and Xinyu Yang and Xiongcai Luo and Xiuli Chen and Xuantong Zhong and Xuefeng Xiao and Xujing Li and Yan Wu and Yawei Wen and Yifan Du and Yihao Zhang and Yining Ye and Yonghui Wu and Yu Liu and Yu Yue and Yufeng Zhou and Yufeng Yuan and Yuhang Xu and Yuhong Yang and Yun Zhang and Yunhao Fang and Yuntao Li and Yurui Ren and Yuwen Xiong and Zehua Hong and Zehua Wang and Zewei Sun and Zeyu Wang and Zhao Cai and Zhaoyue Zha and Zhecheng An and Zhehui Zhao and Zhengzhuo Xu and Zhipeng Chen and Zhiyong Wu and Zhuofan Zheng and Zihao Wang and Zilong Huang and Ziyu Zhu and Zuquan Song},
      year={2025},
      eprint={2505.07062},
      archivePrefix={arXiv},
      primaryClass={cs.CV},
      url={https://arxiv.org/abs/2505.07062}, 
}

@misc{duan2025vlmevalkitopensourcetoolkitevaluating,
      title={VLMEvalKit: An Open-Source Toolkit for Evaluating Large Multi-Modality Models}, 
      author={Haodong Duan and Xinyu Fang and Junming Yang and Xiangyu Zhao and Yuxuan Qiao and Mo Li and Amit Agarwal and Zhe Chen and Lin Chen and Yuan Liu and Yubo Ma and Hailong Sun and Yifan Zhang and Shiyin Lu and Tack Hwa Wong and Weiyun Wang and Peiheng Zhou and Xiaozhe Li and Chaoyou Fu and Junbo Cui and Jixuan Chen and Enxin Song and Song Mao and Shengyuan Ding and Tianhao Liang and Zicheng Zhang and Xiaoyi Dong and Yuhang Zang and Pan Zhang and Jiaqi Wang and Dahua Lin and Kai Chen},
      year={2025},
      eprint={2407.11691},
      archivePrefix={arXiv},
      primaryClass={cs.CV},
      url={https://arxiv.org/abs/2407.11691}, 
}

@misc{zhang2025pelicanvl10foundationbrain,
      title={Pelican-VL 1.0: A Foundation Brain Model for Embodied Intelligence}, 
      author={Yi Zhang and Che Liu and Xiancong Ren and Hanchu Ni and Shuai Zhang and Zeyuan Ding and Jiayu Hu and Hanzhe Shan and Zhenwei Niu and Zhaoyang Liu and Shuang Liu and Yue Zhao and Junbo Qi and Qinfan Zhang and Dengjie Li and Yidong Wang and Jiachen Luo and Yong Dai and Zenglin Xu and Bin Shen and Qifan Wang and Jian Tang and Xiaozhu Ju},
      year={2025},
      eprint={2511.00108},
      archivePrefix={arXiv},
      primaryClass={cs.LG},
      url={https://arxiv.org/abs/2511.00108}, 
}

@misc{wang2025stepsearchignitingllmssearch,
      title={StepSearch: Igniting LLMs Search Ability via Step-Wise Proximal Policy Optimization}, 
      author={Ziliang Wang and Xuhui Zheng and Kang An and Cijun Ouyang and Jialu Cai and Yuhang Wang and Yichao Wu},
      year={2025},
      eprint={2505.15107},
      archivePrefix={arXiv},
      primaryClass={cs.CL},
      url={https://arxiv.org/abs/2505.15107}, 
}

@misc{wang2025eraseimproveerasablereinforcement,
      title={Erase to Improve: Erasable Reinforcement Learning for Search-Augmented LLMs}, 
      author={Ziliang Wang and Kang An and Xuhui Zheng and Faqiang Qian and Weikun Zhang and Cijun Ouyang and Jialu Cai and Yuhang Wang and Yichao Wu},
      year={2025},
      eprint={2510.00861},
      archivePrefix={arXiv},
      primaryClass={cs.CL},
      url={https://arxiv.org/abs/2510.00861}, 
}

%%%%%%%%%%%%%%%%%%%%%%%%%%%%%%%%%%%%%%%%%%%%%%%%%%%%%%%%%%%%

% \appendix

%%%%%%%%%%%%%%%%%%%%%%%%%%%%%%%%%%%%%%%%%%%%%%%%%%%%%%%%%%%%

\end{document}